\documentclass{article}

\usepackage{arxiv}

\usepackage[utf8]{inputenc}
\usepackage[T1]{fontenc}
\usepackage[hidelinks]{hyperref}
\usepackage{url}
\usepackage{booktabs}
\usepackage{amsfonts}
\usepackage{nicefrac}
\usepackage{microtype}
\usepackage{lipsum}
\usepackage{graphicx}
\usepackage[numbers]{natbib}
\usepackage{doi}
\usepackage{siunitx}
\usepackage{subcaption}

\title{Even Faster Simulations with Flow Matching: \\ A Study of Zero Degree Calorimeter Responses}

\author{Maksymilian Wojnar\\
AGH University of Krakow\\
Kraków, Poland\\
\texttt{maksymilian.wojnar@agh.edu.pl}
}

\date{}
\renewcommand{\headeright}{}
\renewcommand{\undertitle}{}
\renewcommand{\shorttitle}{Even Faster Simulations with Flow Matching}

\hypersetup{
pdftitle={Even Faster Simulations with Flow Matching: A Study of Zero Degree Calorimeter Responses},
pdfauthor={Maksymilian Wojnar},
pdfkeywords={fast simulation, flow matching, generative neural network, high-energy physics, zero degree calorimeter},
}

\begin{document}
\maketitle

\begin{abstract}
    Recent advances in generative neural networks, particularly flow matching (FM), have enabled the generation of high-fidelity samples while significantly reducing computational costs. A promising application of these models is accelerating simulations in high-energy physics (HEP), helping research institutions meet their increasing computational demands. In this work, we leverage FM to develop surrogate models for fast simulations of zero degree calorimeters in the ALICE experiment. We present an effective training strategy that enables the training of fast generative models with an exceptionally low number of parameters. This approach achieves state-of-the-art simulation fidelity for both neutron (ZN) and proton (ZP) detectors, while offering substantial reductions in computational costs compared to existing methods. Our FM model achieves a Wasserstein distance of 1.27 for the ZN simulation with an inference time of \SI{0.46}{\milli\second} per sample, compared to the current best of 1.20 with an inference time of approximately \SI{109}{\milli\second}. The latent FM model further improves the inference speed, reducing the sampling time to \SI{0.026}{\milli\second} per sample, with a minimal trade-off in accuracy. Similarly, our approach achieves a Wasserstein distance of 1.30 for the ZP simulation, outperforming the current best of 2.08. The source code is available at \url{https://github.com/m-wojnar/faster_zdc}.
\end{abstract}

\keywords{fast simulation \and flow matching \and generative neural network \and high-energy physics \and zero degree calorimeter}

\section{Introduction}
\label{sec:introduction}

Simulation, alongside analytical modeling and experimentation, is a foundational element of contemporary scientific research. The need for specialized simulations is steadily increasing across numerous areas of science and technology. A prime example of an institution reliant on complex multi-scale simulations is CERN, the leading center for high-energy physics (HEP), which employs high-performance computing to predict detector responses, handle massive real-time data streams, and store the extensive data volumes acquired during LHC (Large Hadron Collider) runs. As CERN expands, so does the demand for computational power and data storage. With the forthcoming initiation of the HL-LHC (High-Luminosity LHC) planned for 2029, storage and processing demands are projected to increase by an order of magnitude compared to present levels~\cite{cern2025computing}.

To address the rising computational demands imposed by classical HEP simulation models that rely on Monte Carlo methods, there is an increasing need for surrogate models. These models provide approximate simulation outcomes with significantly reduced computational expenses. In its strategic planning for the coming years, CERN identifies the need for fast simulation models as a critical element of advancement~\cite{cern2019roadmap}. A well-recognized approach for developing surrogate models is leveraging machine learning (ML) techniques, especially by employing generative neural networks~\cite{apostolakis2020}.

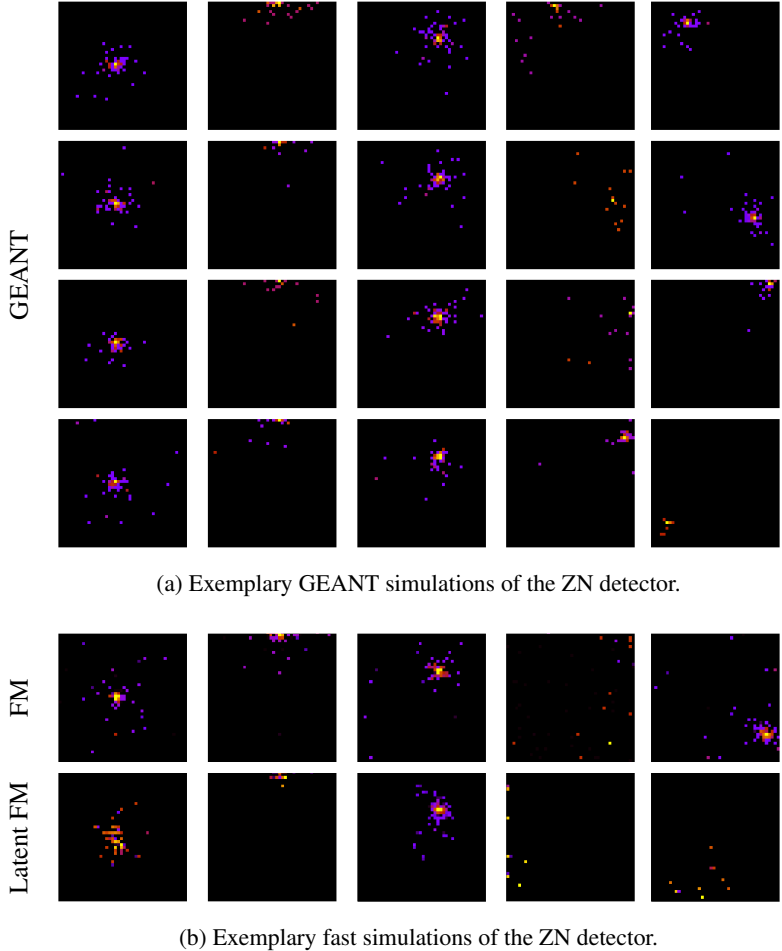
\begin{figure}[t!]
    \centering
    \begin{subfigure}[t]{0.6\linewidth}
        \centering
        \includegraphics[width=\linewidth]{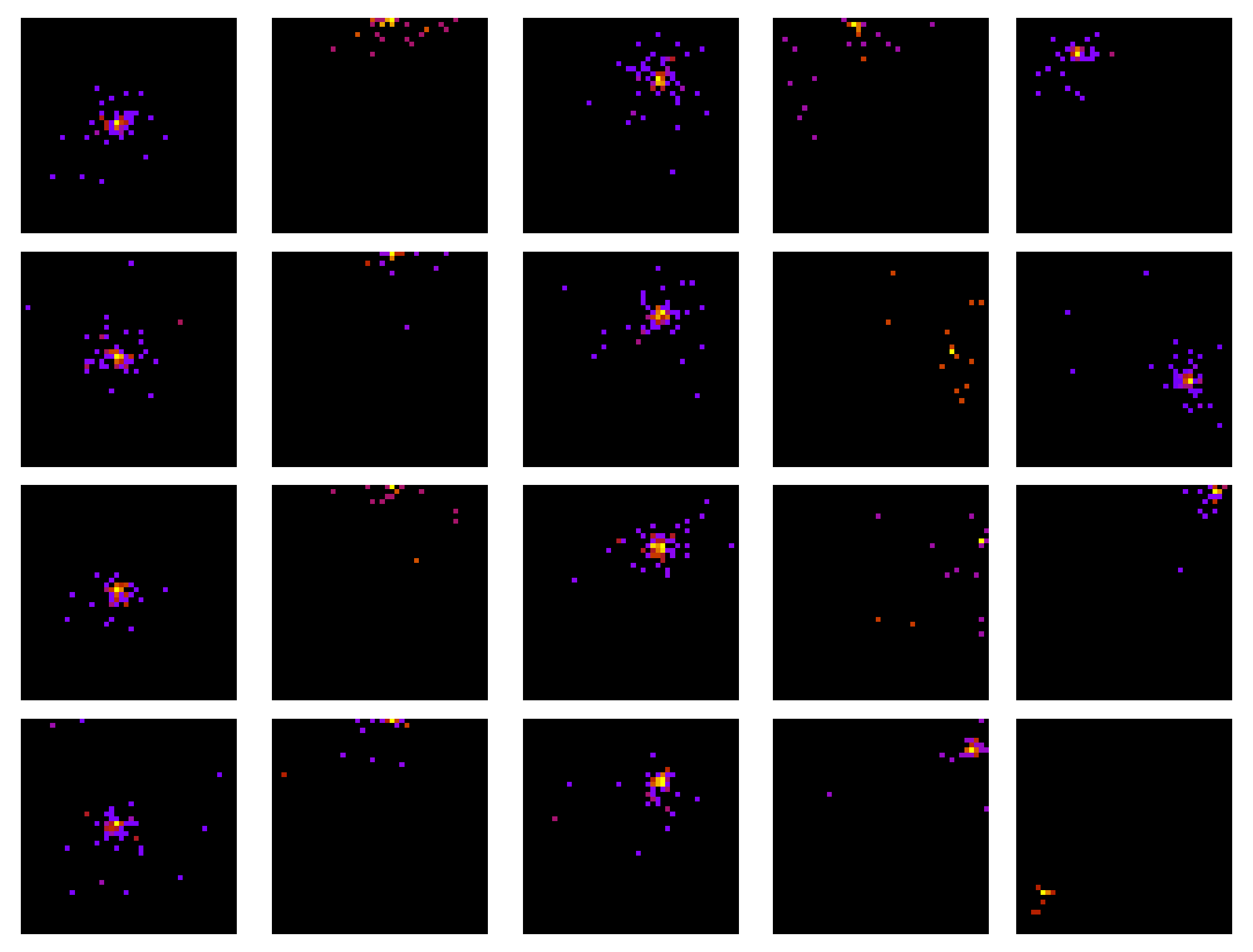}
        \begin{picture}(0,0)
            \put(-150, 118){\makebox(0,0){\rotatebox{90}{GEANT}}}
        \end{picture}
        \vspace{-4mm}
        \caption{Exemplary GEANT simulations of the ZN detector.}
        \vspace{3mm}
        \label{fig:sample_n}
    \end{subfigure}
    \\
    \begin{subfigure}[t]{0.6\linewidth}
        \centering
        \includegraphics[width=\linewidth]{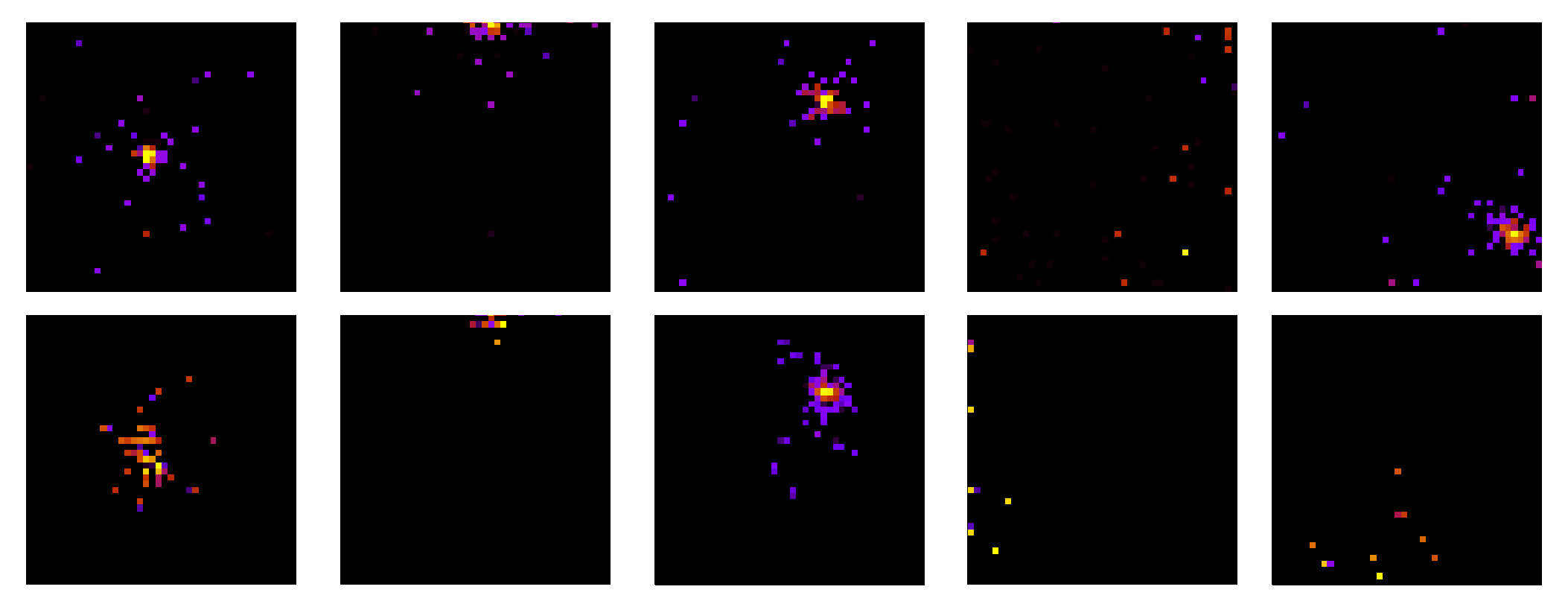}
        \begin{picture}(0,0)
            \put(-150, 92){\makebox(0,0){\rotatebox{90}{FM}}}
            \put(-150, 39){\makebox(0,0){\rotatebox{90}{Latent FM}}}
        \end{picture}
        \vspace{-4mm}
        \caption{Exemplary fast simulations of the ZN detector.}
        \label{fig:sample_fast_n}
    \end{subfigure}
    \caption{Example ZN responses generated with GEANT and FM. The columns correspond to the same particles, while the rows are the results of independent simulations. The first three columns on the left correspond to particles with low response diversity, while the two on the right to particles with high diversity.}
    \label{fig:samples}
\end{figure}

An active area of research focuses on employing generative neural networks for fast simulation of zero degree calorimeters (ZDC) within ALICE (A Large Ion Collider Experiment)~\cite{gallio1999zdc}, which is simulated using the resource-intensive GEANT Monte Carlo toolkit~\cite{geant}. The ZDC detectors are located \SI{112.5}{\meter} from the interaction point and require simulating particles through several layers of matter. Additionally, the experimental setup includes two types of detectors -- a proton detector (ZP) with a resolution of $56 \times 30$ and a neutron detector (ZN) with a resolution of $44 \times 44$ -- arranged in pairs on either side of the ALICE detector system. The detectors, measuring the Cherenkov radiation, count photons from bundles of optical fibers divided into five channels -- one for each quadrant and one from the entire detector area, with each channel connected to every second optical fiber of the detector. Therefore, the outputs of the ALICE electronic devices are five values corresponding to the sums of photons in each channel. However, computer simulation can provide more precise predictions by calculating values for each individual fiber. Consequently, GEANT (as well as the target surrogate model) generates a two-dimensional array of photon counts. Simulating ZDC responses is particularly challenging, as the detector exhibits varying response diversity to different particles. For some particles, the detector response remains consistent across independent runs, whereas for others, the resulting particle shower varies, even when the input parameters are identical (Figure~\ref{fig:samples}).

\begin{figure}[t!]
    \centering
    \begin{subfigure}[t]{0.61\linewidth}
        \centering
        \includegraphics[width=\linewidth]{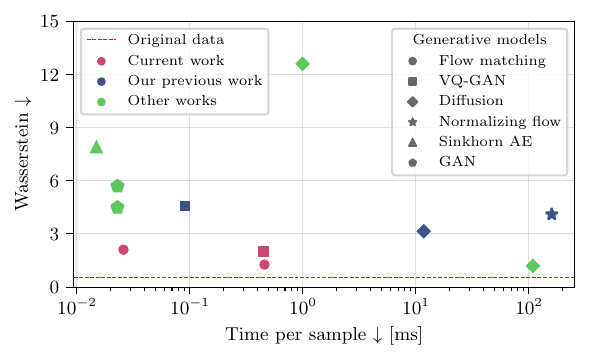}
        \caption{ZDC neutron detector.}
        \label{fig:all_n}
    \end{subfigure}
    \hfill
    \begin{subfigure}[t]{0.3\linewidth}
        \centering
        \includegraphics[width=\linewidth]{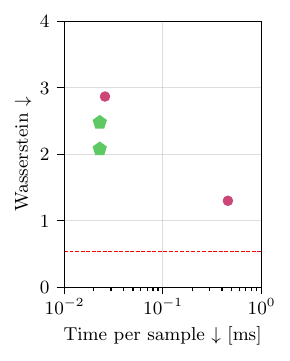}
        \caption{ZDC proton detector.}
        \label{fig:all_p}
    \end{subfigure}
    \caption{Comparison of the Wasserstein distance and sampling time in neutron (left) and proton (right) detector simulations, highlighting differences between this study, previous work~\cite{wojnar2025fast, wojnar2024applying}, and other research. Note the logarithmic scale on the x-axis.}
    \label{fig:all}
\end{figure}

The task of fast ZDC simulation is well documented in the literature, with numerous models proposed that address challenges related to this task. These challenges include faithful reproduction of the particle shower characteristics, precise simulation of the underlying physical phenomena, varying response diversity, and significant dataset imbalance (Appendix~\ref{sec:dataset}). However, several issues continue to hinder the practical implementation of these solutions. Generative adversarial networks (GANs) and autoencoders tend to be fast but less accurate, while diffusion models offer greater precision but require more computational resources~\cite{wojnar2025fast}. Furthermore, up to now, just one research effort has concentrated on simulating a ZP detector. To address these gaps, we propose a surrogate model based on flow matching (FM). This work builds on the research presented in~\cite{wojnar2025fast} and~\cite{wojnar2024applying}, employing a comparable methodology. However, it introduces a novel approach for fast and accurate ZDC simulation, making the following key contributions:

\begin{itemize}
    \item We propose an FM model specifically designed for ZDC, featuring a remarkably low parameter count (in the tens of thousands), significantly fewer than currently used large generative models (Section~\ref{sec:flow_matching}).
    \item Our FM model achieves superior performance in the simulation of ZN and ZP detector responses, ensuring high-fidelity results (Section~\ref{sec:results}).
    \item The proposed latent FM approach sets a new benchmark in simulation speed, significantly reducing computational costs compared to existing methods (Figure~\ref{fig:all}).
    \item We improve the VQ-GAN model for the ZN detector compared to previous work~\cite{wojnar2025fast, wojnar2024applying} and present new promising results (Appendix~\ref{sec:vq_gan}).
    \item This work is the first to address both the simulation and data analysis of ZN and ZP detector responses (Appendix~\ref{sec:dataset}).
\end{itemize}

\newpage
\section{Related Work}
\label{sec:related_work}

Generative models for fast detector simulation are explored at CERN from 2017. The LAGAN model~\cite{deOliveira2017} introduces GANs adapted to sparse data, where spatial location plays a key role. In 2018, 3D GAN~\cite{khattak2018three} extends this idea to the Compact Linear Collider detector, showing the potential of GANs for complex 3D geometries. In 2020, a modified variational autoencoder (VAE)~\cite{dohi2020variational} is used to generate diverse particle physics events with control in a continuous latent space (based on the dataset from~\cite{deOliveira2017}). Presented a year later, CaloFlow~\cite{krause2021caloflow} applies normalizing flows (NFs) to calorimeter simulations. Its successor, CaloFlow II~\cite{krause2021caloflowii} further improves performance, achieving speeds nearly $10^4$ times faster than GEANT.

The 2022 Calo Challenge~\cite{calochallenge2022} highlights the need for efficient simulation tools and provides a common dataset for model development. CaloMan~\cite{cresswell2022CaloMan} responds with a generalized autoencoder that reduces dimensionality, speeding up training and generation. That same year, the first diffusion model for calorimeter simulation is presented~\cite{mikuni2022score}. In 2023, a method combining autoencoders with NF~\cite{orzari2023lhc} enables analytical sampling in latent space. The authors of~\cite{cresswell2024scaling} show in their 2024 article that XGBoost~\cite{chen2016xgboost} can be used as a function approximator in diffusion and FM models on simplified calorimeter datasets. Finally, CaloDREAM~\cite{favaro2025calodream} from 2025 combines FM with transformers to simulate detector responses more accurately, marking one of the first uses of FM in this domain.

The first study to use ML for fast ZDC simulation from 2021 introduces the end-to-end Sinkhorn Autoencoder~\cite{deja2021end}. This architecture provides a more accurate representation of detector responses and better preservation of physical properties (e.g., improved placement of collision centers) compared to standard VAE and GAN baselines. However, it is still insufficient to realistically reproduce the characteristics of particle showers, which appear as scattered flashes. GANs closely resemble the results of Monte Carlo simulations in terms of visual appearance, but they fail to preserve physical properties and handle the variability of particle interactions. 

To address the diversity issue, SDI-GAN~\cite{dubinski2023selectively} has been introduced in 2023. The proposed method selectively enhances the diversity of GAN-generated samples by incorporating an additional regularization factor into the training loss function. In their 2024 article, the authors of~\cite{dubinski2024machine} compare VAE, GAN, and GAN with proposed improvements. These include an auxiliary network for identifying zero detector responses (whereas other studies assume simulations only for collisions producing a minimum number of photons) and regularization techniques that enhance accuracy of the predicted particle interaction centers.

Other studies from 2024 use CorrVAE to encode response characteristics across different dimensions of the latent space~\cite{rogozinski2024particle}. The authors of~\cite{kita2024generative} employ generative diffusion models for ZN simulation and establish a new benchmark in simulation fidelity. However, its long generation time poses a significant challenge, effectively limiting its applicability for fast ZDC simulations. In~\cite{wojnar2024applying}, later expanded in~\cite{wojnar2025fast} in 2025, an extensive comparison of generative neural networks is presented. The study introduce the use of new architectures as surrogate models and compare previously applied autoencoders, GANs, and diffusion models with VQ-VAE (vector quantized variational autoencoder), VQ-GAN, and NF. In the fast ZDC simulation task, VQ techniques outperform traditional autoencoders and GANs, offering a compromise between the inference time and the simulation fidelity. On the other hand, despite even higher accuracy of diffusion models and NFs, they are constrained by longer generation durations.

The first results on ZP simulation, reported in 2024 in~\cite{bedkowski2024deep}, use SDI-GAN along with additional regularization techniques that enhance simulation fidelity. However, this remains the only study on the topic, highlighting a significant gap in the research. Furthermore, state-of-the-art models have not yet been thoroughly evaluated, leaving their performance and applicability untested.

\section{Background}
\label{sec:background}

The key aspects of the proposed FM-based approach are outlined below. Appendix~\ref{sec:setup} provides further implementation details and comprehensive training guidelines for FM and latent FM models.

\subsection{Metrics}
\label{sec:metrics}

The first step in assessing the quality of a surrogate model for fast ZDC simulation is to define the evaluation metrics. Since the characteristics of particle showers are inherently stochastic and the detector responses exhibit scattered burst-like patterns, classical image comparison methods (e.g., pixel-wise mean squared error) are not suitable measures of accuracy. Instead, the evaluation relies on comparing the channel values described in Section~\ref{sec:introduction}, which correspond to the electronic outputs of the devices installed in ALICE. 

There are two primary approaches in the literature. The first involves computing the average Wasserstein distance between the histograms of the original and generated data. Note that these histograms, representing the photon counts in a given channel, are calculated across the entire dataset and are presented in Appendix~\ref{sec:histograms}. Formally, this metric can be defined as
\begin{equation}
    \text{Wasserstein-1}(w, \hat{w}) = \frac{1}{m} \sum_{i=1}^m  \int_0^1 \left| F^{-1}_{w_i}(z) - F^{-1}_{\hat{w}_i}(z) \right| dz,
\end{equation}
where $m$ is the number of channels, $F^{-1}_q$ is the inverse cumulative distribution function of the distribution $q$, $w_i$ denotes the true distribution of the $i$-th channel, and $\hat{w}_i$ is the distribution of the predicted values. The Wasserstein distance is widely used in detector calibration for comparing marginal distributions due to the stochastic nature of detector responses. Although coarse-grained, it remains the most informative and domain-relevant metric.

The second metric is the mean absolute error (MAE), which evaluates the differences between channel values on a per-example basis. Unlike the first metric, where the aim is to minimize its value, the MAE metric goal is to closely approximate values observed in the original data. This arises from the stochastic nature of showers, where multiple detector responses may be valid for a given particle. Specifically, a high degree of similarity between individual examples is desirable in low-diversity regimes, while greater variability is expected for high-diversity particles. The MAE metric is defined as
\begin{equation}
    \text{MAE}(w, \hat{w}) = \frac{1}{n} \sum_{k=1}^n \sum_{i=1}^m |w_i^k - \hat{w}_i^k|,
\end{equation}
where $n$ refers to the number of evaluated examples, $w_i^k$ represents the value of the $i$-th channel of the $k$-th response. Due to the randomness of the process, the reported values are the average results of five runs for the entire test set.

The Wasserstein metric for the original dataset is computed by randomly splitting the test set into two equal parts, with one serving as detector responses and the other as generated samples. The same approach is used for the MAE metric, but with the additional requirement that the compared examples share identical particle features.

\subsection{Flow Matching}
\label{sec:flow_matching}

FM is a family of generative models that facilitate the transition from the noise $x_0$ to data $x_1$ through a linear interpolation process
\begin{equation}
    x_t = (1 - t)x_0 + tx_1,
\end{equation}
where $t \in [0, 1]$ is the interpolation time~\cite{lipman2023flow, liu2022flow, albergo2023build}. We follow~\cite{lipman2023flow} and train the neural network to learn the normalized transition velocity
\begin{equation}
    v_t = x_1 - x_0,
\end{equation}
where $x_0$ is the noise sampled at time $t$. Then the velocity can be used to generate samples incrementally by applying the Euler method
\begin{equation}
    x_{t+\Delta t} = x_t + \Delta t \cdot v_t.
\end{equation}
Due to the intractability of the loss function, FM is not used in practice~\cite{lipman2023flow}. Instead, conditional flow matching is adopted with a modified, tractable loss function and simplified sampling defined as
\begin{equation}
    \mathcal{L}(\theta) = \mathbb{E}_{t, q(x_1), p(x_0)} ||v_\theta(x_t, t)-v_t||^2,
\end{equation}
where $q$ is a data distribution, $p$ is the Gaussian distribution, and $v_\theta(x_t, t)$ represents the velocity predicted by a neural network parametrized by $\theta$ with input $x_t$ and time $t$ sampled from the uniform distribution.

\section{Methodology}
\label{sec:training}

\begin{figure}[t!]
    \centering
    \includegraphics[width=0.35\linewidth]{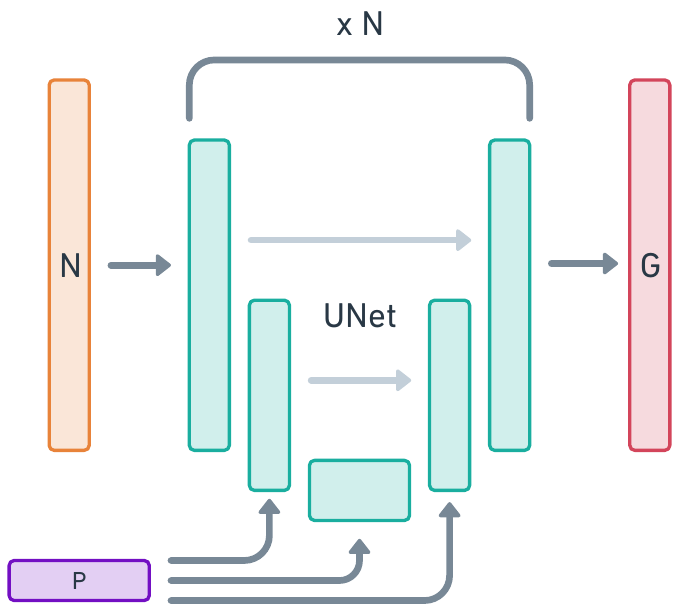}
    \caption{Schematic illustration of the FM model. The input noise $x_0$ is shown in orange (N), the generated sample $x_1$ in red (G), and the conditional variables representing the particle features in purple (P). The neural network, based on the U-Net architecture, is shown in green and is applied multiple times to incrementally generate a sample.}
    \label{fig:fm_unet}
\end{figure}

The architecture of the neural network is based on the Stable Diffusion autoencoder~\cite{rombach2022high} and includes an attention mechanism enabling conditioning on the input vector. For this study, we implement a compact U-Net~\cite{ronneberger2015unet} with 77k parameters and a linear noise schedule (Figure~\ref{fig:fm_unet}). We apply additional improvements to speed up the model and adjust the number of steps (Appendix~\ref{sec:setup}).

The training process of the latent FM model consists of two stages. First, a small VAE is trained using the same architecture with a downsize factor of 4. To ensure that the reconstructed samples closely resemble the original GEANT data, the VAE training employs a multi-component loss function and gradient normalization. Subsequently, the FM training procedure is repeated in the latent space. A detailed description of this process can be found in Appendix~\ref{sec:setup}.

\begin{figure}[t!]
    \centering
    \includegraphics[width=0.6\linewidth]{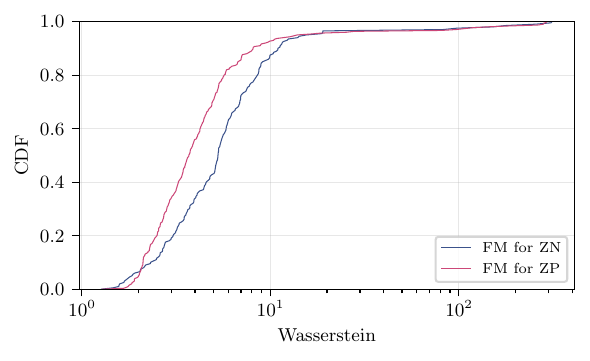}
    \caption{CDF of the Wasserstein distance over 300 hyperparameter optimization trials. The curves indicate that highly performant runs (Wasserstein $< 2$) occur in less than 10\% cases.}
    \label{fig:cdf}
\end{figure}

Due to the challenges associated with dataset characteristics, as well as the significant reduction in both the number of layers and model parameters compared to previous works, network training does not always achieve convergence. Figure~\ref{fig:cdf} presents the cumulative distribution function (CDF) of the Wasserstein distance, illustrating the frequency of successful optimization outcomes. In our experiments with FM models for the ZN detector, the value of 1.4 is achieved after 49 trials. To address the instability issue, multiple training runs and hyperparameter tuning of the Adam optimizer~\cite{kingma2017adam} are performed using the Optuna framework~\cite{akiba2019optuna}. The compact size of the network and the use of NVIDIA A100 GPUs with 40 GB of memory installed in the Athena supercomputer~\cite{athena} allows for the rapid training process and the evaluation of hundreds of configurations.

\section{Results}
\label{sec:results}

Tables~\ref{tab:table_n} and~\ref{tab:table_p} provide the performance comparison of various generative models in simulating the ZN and ZP detector responses, respectively. In the ZN simulation task, FM achieves a Wasserstein distance of 1.27 and MAE of 16.99, closely matching the best-performing diffusion model (Wasserstein: 1.20) but at a significantly lower computational cost (\SI{0.46}{\milli\second} vs. \SI{109}{\milli\second} per sample). Furthermore, Latent FM offers an even faster inference time of \SI{0.026}{\milli\second}, rivaling GAN-based approaches while maintaining reasonable accuracy. VQ-GAN and diffusion demonstrate a trade-off between sample quality and inference speed. Similarly, in the ZP simulation task, FM outperforms existing models, achieving a Wasserstein distance of 1.30 and an MAE of 13.69 (Table~\ref{tab:table_p}). Latent FM provides a faster alternative, albeit with a slight performance trade-off. The same results are visualized in Figure~\ref{fig:all} for improved interpretability.

\begin{table}[t!]
    \small
	\centering
	\caption{Performance comparison of generative models in the ZN simulation task. The table reports the inference time per sample with batch size of 256. The asterisk ($^*$) denotes a diffusion model trained using a different pipeline and hardware~\cite{kita2024generative}, accounting for the absence of MAE results and the potential incomparability of sampling time.}
	\begin{tabular}{lccc}
		\toprule
		\textbf{Model}     & \textbf{Wasserstein} $\downarrow$ & \textbf{MAE} & \textbf{Time [ms]} $\downarrow$ \\
        \midrule
        Original data      & 0.53          & 16.41          & --             \\
        \specialrule{0.4pt}{0pt}{0pt}
        GAN                & 5.70          & 24.71          & \textbf{0.023} \\
        NFs                & 4.11          & 19.36          & 160.00         \\
        Diffusion          & 3.15          & 20.10          & 11.80          \\
        VQ-GAN             & 2.01          & 20.33          & 0.45           \\
        Diffusion$^*$      & \textbf{1.20} & --             & 109.00         \\
        \specialrule{0.4pt}{0pt}{0pt}
        Latent FM          & 2.11          & 22.32          & \textbf{0.026} \\
        FM                 & \textbf{1.27} & \textbf{16.99} & 0.46           \\
        \bottomrule
	\end{tabular}
	\label{tab:table_n}
\end{table}

\begin{figure}[t!]
    \centering
    \includegraphics[width=0.6\linewidth]{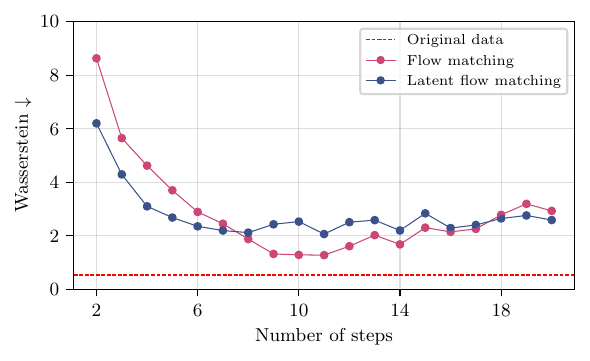}
    \caption{Performance of FM and latent FM as a function of the number of sampling steps.}
    \label{fig:steps}
\end{figure}

\begin{table}[t!]
    \small
	\centering
	\caption{Performance comparison of generative models in the ZP simulation task. The table reports the inference time per sample with batch size of 256. The asterisk ($^*$) denotes GAN models trained using a different pipeline~\cite{bedkowski2024deep}, accounting for the absence of MAE results and the potential incomparability of sampling time.}
	\begin{tabular}{lccc}
		\toprule
		\textbf{Model}     & \textbf{Wasserstein} $\downarrow$ & \textbf{MAE} & \textbf{Time [ms]} $\downarrow$ \\
        \midrule
        Original data      & 0.56          & 13.00          & --             \\
        \specialrule{0.4pt}{0pt}{0pt}
        GAN$^*$            & 2.48          & --             & \textbf{0.023} \\
        SDI-GAN$^*$        & 2.08          & --             & \textbf{0.023} \\
        \specialrule{0.4pt}{0pt}{0pt}
        Latent FM          & 2.87          & 15.64          & \textbf{0.026} \\
        FM                 & \textbf{1.30} & \textbf{13.69} & 0.46           \\
        \bottomrule
	\end{tabular}
	\label{tab:table_p}
\end{table}

Figure~\ref{fig:steps} illustrates the simulation fidelity for both FM and latent FM models relative to the number of steps in the Euler method (Section~\ref{sec:flow_matching}). In each case, a distinct minimum is observed around 10 steps, after which the quality of generation declines. This phenomenon might result from the initial setting of 10 steps during hyperparameter tuning described in Appendix~\ref{sec:setup}. Eventually, 11 steps are selected for the FM model and 7 for the latent FM model, as these proved to offer the best performance.

The performance optimizations applied to improve the inference time of the FM model are summarized in Table~\ref{tab:time_fm}. The original model, adopted from~\cite{wojnar2025fast} and sharing the same architecture as the diffusion model, has an inference time of 11.80 ms. Reducing the number of steps -- primarily achieved by transitioning from the diffusion to FM framework -- leads to a significant performance improvement, reducing the inference time by 78\%. Additionally, using a smaller model along with a mixed precision training and inference with the \textit{bfloat16} data type further reduced the inference time to \SI{0.46}{\milli\second}. The most substantial performance gain is achieved through the introduction of the latent FM model, resulting in a final inference time of \SI{0.026}{\milli\second} per sample.

\begin{table}[t!]
    \small
    \centering
    \caption{Performance optimizations and their impact on FM model inference time (for batch size of 256). The last column highlights the improvement gained by applying optimizations sequentially, relative to the previous one.}
    \begin{tabular}{lcc}
        \toprule
        \textbf{Improvement}         & \textbf{Time [ms]} $\downarrow$ & \textbf{Relative change} \\
        \midrule
        Base model (diffusion from~\cite{wojnar2025fast}) & 11.80                & --                       \\
        \specialrule{0.4pt}{0pt}{0pt}
        Fewer number of steps (50 $\rightarrow$ 11)   & 2.60                 & -78\%                    \\
        Smaller model (4M $\rightarrow$ 77k params)   & 0.62                 & -76\%                    \\
        Mixed precision (F32 $\rightarrow$ F16)        & 0.46                 & -26\%                    \\
        Generation in a latent space & 0.026                & -94\%                    \\
        \bottomrule
    \end{tabular}
	\label{tab:time_fm}
\end{table}

\section{Discussion and Practical Applications}
\label{sec:discussion}

By significantly reducing inference time while preserving high-quality detector response simulation, our method sets a new benchmark in the speed and fidelity of ZDC modeling, making it a strong candidate for large-scale simulations at CERN, especially in preparation for the HL-LHC era. The ALICE experiment at CERN already supports running ML models through ONNX Runtime~\cite{onnxruntime}, enabling efficient inference across different hardware architectures~\cite{aliceML}. Our FM model, implemented in JAX~\cite{jax2018github}, can be easily converted to the ONNX format, making it compatible with ALICE’s existing ML infrastructure. The source code, hyperparameters, weights, and ONNX checkpoints are available in the accompanying repository.

Despite its advantages, FM also presents some challenges. Training stability remains an issue, as models are sensitive to hyperparameter choices and require careful tuning to avoid divergence. Moreover, the procedure of training latent FM is intricate, involving a complex loss function and regularization (Appendix~\ref{sec:setup}), requiring specialized expertise. Additionally, while FM offers a strong balance between fidelity and efficiency, the trade-off is evident in latent FM. This approach significantly reduces inference time at the cost of lower accuracy, making it more suitable for applications where speed is prioritized over precision.

Future work will focus on creating a larger and more balanced dataset that accurately covers the space of primary particles. This task is demanding due to the extensive computational resources required and the complexity of the scientific software involved. Moreover, approaches that enhance convergence and simplify model training would be highly beneficial. Investigation of alternative strategies for developing smaller, optimized FM models is also very promising, e.g., knowledge distillation or parameter pruning followed by fine-tuning.

\section{Acknowledgments}

We would like to thank Emilia Majerz, Professor Witold Dzwinel, and Professor Jacek Kitowski from AGH University of Krakow, Professor Jacek Otwinowski from the Institute of Nuclear Physics PAN in Krakow, and Sandro Wenzel, PhD from CERN. 

This work is in part supported by the Ministry of Science and Higher Education (Agreement Nr 2023/WK/07) by the program entitled ``PMW'' and by the Ministry funds assigned to AGH University in Krakow. We gratefully acknowledge Polish high-performance computing infrastructure PLGrid (HPC Center: ACK Cyfronet AGH) for providing computer facilities and support within computational grant no. PLG/2024/017264.

\clearpage
\appendix

\section{Dataset Analysis}
\label{sec:dataset}

\begin{figure}[t!]
    \centering
    \begin{subfigure}[t]{\linewidth}
        \centering
        \includegraphics[width=0.4\linewidth]{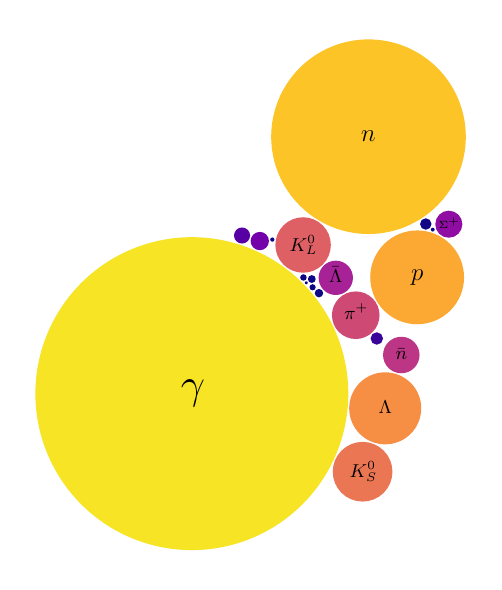}
        \caption{Particle type distribution in the ZN response dataset: $\gamma$ (60.83\%), $n$ (23.34\%), $p$ (5.30\%), $\Lambda$ (3.08\%), $K_S^0$ (2.09\%), $K_L^0$ (1.79\%), $\pi^{+}$ (1.30\%), with all other particle types contributing less than 1\% each.}
        \label{fig:bubble_n}
    \end{subfigure}
    \\
    \begin{subfigure}[t]{\linewidth}
        \centering
        \includegraphics[width=0.4\linewidth]{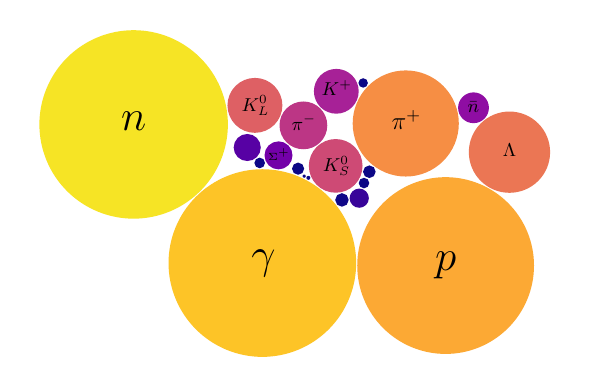}
        \caption{Particle type distribution in the ZP response dataset: $n$ (26.92\%), $\gamma$ (26.67\%), $p$ (23.64\%), $\pi^{+}$ (8.38\%), $\Lambda$ (4.86\%), $K_L^0$ (2.16\%), $K_S^0$ (2.06\%), $\pi^{-}$ (1.62\%), $K^{+}$ (1.41\%), with all other particle types contributing less than 1\% each.}
        \label{fig:bubble_p}
    \end{subfigure}
    \caption{Diagrams illustrating distributions of particle types within the dataset, where $\gamma$ is photon, $n$ is neutron, $p$ is proton, $\Lambda$ is lambda baryon, $K_S^0$ is short-lived neutral kaon, $K_L^0$ is long-lived neutral kaon, $\pi^{+}$ is positive pion, $\bar{n}$ is antineutron, $\bar{\Lambda}$ is anti-lambda baryon, $\Sigma^{+}$ is positive sigma baryon, $\pi^{-}$ is negative pion, $K^{+}$ is positive Kaon.}
    \label{fig:bubbles}
\end{figure}

The dataset generated using the GEANT simulation toolkit includes 21 different particle types, when considering only responses with at least 10 photons. An analysis of the total and unique particle counts (Table~\ref{tab:num_of_particles}) suggests that the dataset may poorly cover the full range of possible particles. Furthermore, a significant class imbalance is evident (Figure~\ref{fig:bubbles}). These limitations could affect the results, potentially leading to lower-quality model predictions for underrepresented particles. Unfortunately, conventional techniques for handling imbalanced datasets are not applicable in this case due to the extreme level of imbalance.

\begin{table}[t!]
    \small
    \centering
    \caption{Summary of the total number of examples and unique particles for the ZN and ZP datasets.}
	\label{tab:num_of_particles}
    \begin{tabular}{lrr}
        \toprule
        \textbf{Statistics}        & \textbf{ZN}         & \textbf{ZP}         \\
        \midrule
        Total number of examples   & \SI{306780}{}     & \SI{129676}{}     \\
        Number of unique particles & \SI{1805}{}       & \SI{2027}{}       \\
        \bottomrule
    \end{tabular}
\end{table}

\begin{figure}[t!]
    \centering
    \begin{subfigure}[t]{0.48\linewidth}
        \centering
        \includegraphics[width=\linewidth]{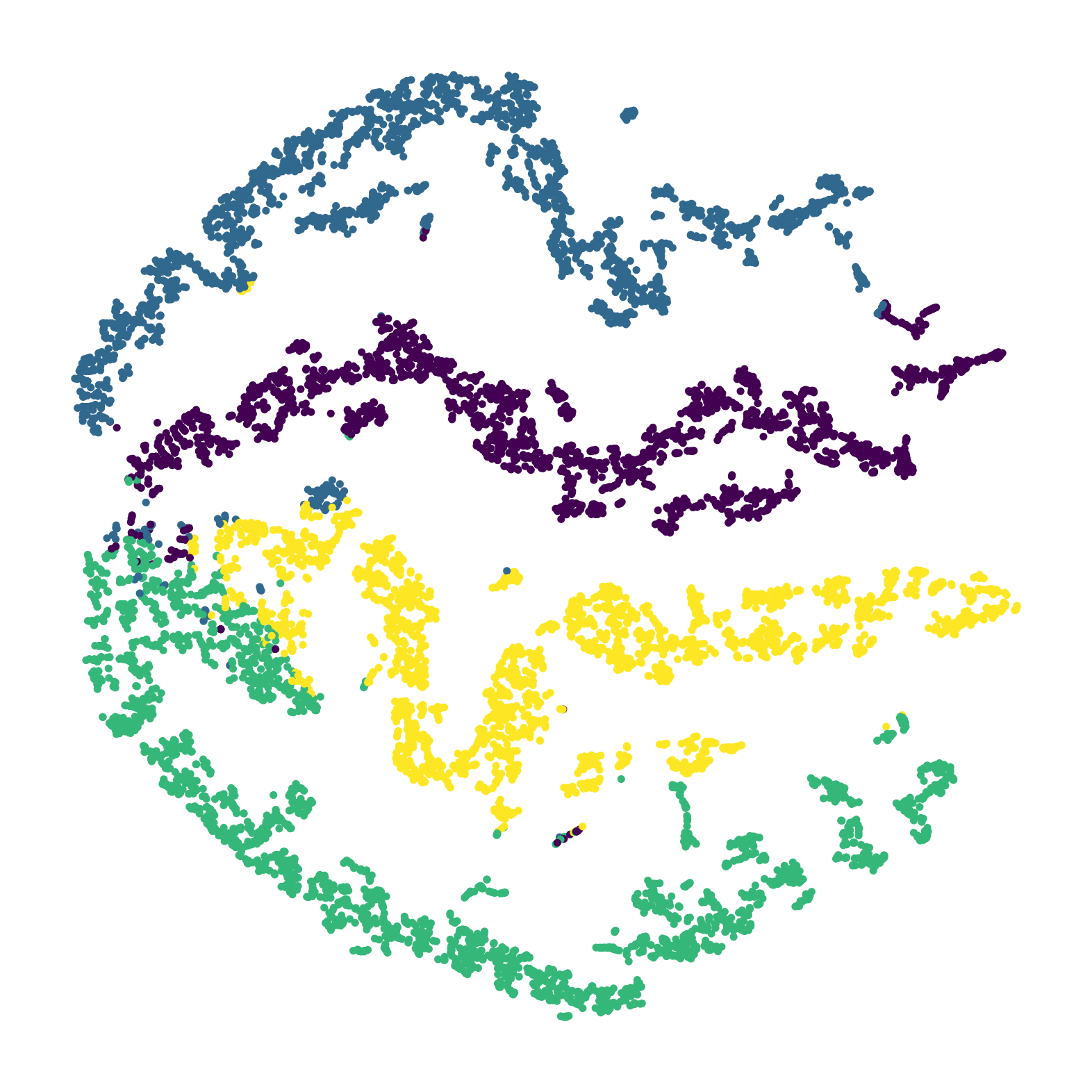}
        \caption{Visualization of the ZN detector dataset.}
        \label{fig:vis_n}
    \end{subfigure}
    \hfill
    \begin{subfigure}[t]{0.48\linewidth}
        \centering
        \includegraphics[width=\linewidth]{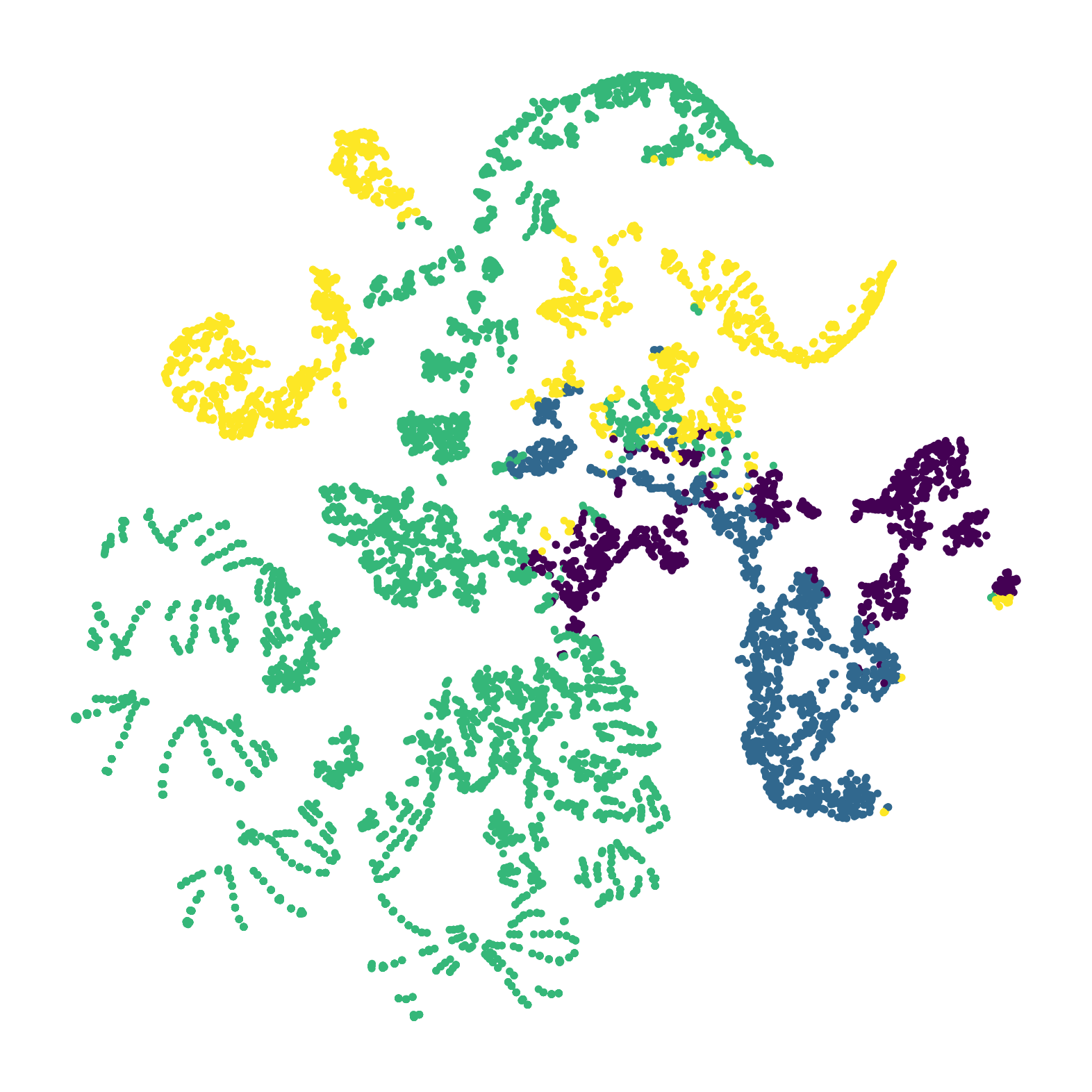}
        \caption{Visualization of the ZP detector dataset.}
        \label{fig:vis_p}
    \end{subfigure}
    \caption{Visualization of particle feature vectors using t-SNE. Colors represent the channel number with the highest photon sum (considering only the first four channels, i.e., quadrants).}
    \label{fig:vis}
\end{figure}

Visualizations of the particle feature space are generated using t-distributed stochastic neighbor embedding (t-SNE)~\cite{maaten2008visualizing}, with \SI{10000}{} examples randomly sampled from the dataset and normalized features (Figure~\ref{fig:vis}). Colors indicate the approximate location of the activation center. The embedding more effectively separated ZN particles, which may suggest a more consistent distribution of activation centers depending on the particle features.

\begin{figure}[t!]
    \centering
    \includegraphics[width=0.9\linewidth]{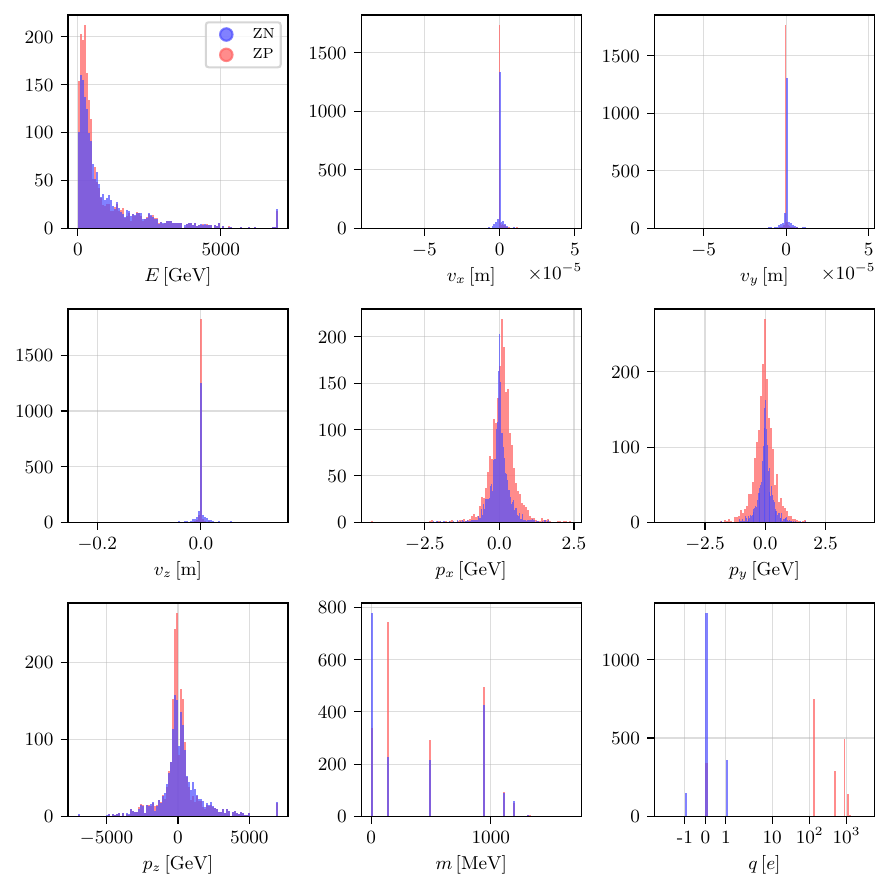}
    \caption{Histograms of particle features for the ZN and ZP detectors. $E$ stands for energy, $v$ for primary vertex positions, $p$ for momenta, $m$ for mass, and $q$ for charge. Note that the x-axis for the $q$ values is in a \textit{symlog} scale to enhance readability.}
    \label{fig:analysis}
\end{figure}

Figure~\ref{fig:analysis} displays histograms of particle features for the ZN and ZP response datasets (i.e., energy, primary vertex positions in 3D, momentum in 3D, mass, and charge). While momentum and mass are conventionally expressed in units of $\frac{\text{GeV}}{c}$ and $\frac{\text{GeV}}{c^2}$, respectively (where $c$ is the speed of light in a vacuum), the values in this dataset are normalized to GeV for simplicity. Additionally, $m$ and $q$ are discrete (quantized) variables.

\section{Experimental Setup}
\label{sec:setup}

The datasets are randomly partitioned into training (70\%), validation (10\%), and testing (20\%) subsets. Conditional variables are standardized to have a mean of zero and a unit variance. To enhance compatibility with a neural network, the ZDC responses are transformed using a logarithmic scale. Performance metrics are computed after applying an inverse transformation to restore the data to its original scale. All models are trained for 50 epochs with a batch size of 256. The generation time is measured with the same batch size, excluding the JAX compilation time.

As discussed in Section~\ref{sec:training}, network training can be unstable, with model quality being highly sensitive to optimizer parameters. Thanks to the small size of the models and the availability of computational resources, an extensive hyperparameter tuning process is conducted, involving at least 300 training trials per model. Training of one FM model takes approximately \SI{25}{\minute} using the Athena supercomputer~\cite{athena}, so the recommended number of trials requires approximately \SI{125}{\hour} of GPU processing time. Optimization is performed using Optuna, with the primary objective of minimizing the Wasserstein distance on the validation dataset. The tuned hyperparameters include the learning rate, $\beta_1$, $\beta_2$, and the application of cosine decay, which have been identified as the most influential factors in~\cite{wojnar2024applying}. After each trial, model weights are saved, and the best-performing model is selected for evaluation on the test set.

The VAE used in this work is based on the architecture proposed by~\cite{rombach2022high} and incorporates a training procedure inspired by~\cite{ryu2024training}. The model has a total of 60k parameters and is designed to ensure high reconstruction fidelity achieved through a loss function including gradient normalization with respect to the input:
\begin{equation}
    \label{eq:loss}
    \mathcal{L}(\theta) = \mathbb{E}_{q(x)} \left[ \frac{\mathcal{L}_{VAE}}{||\nabla_x\mathcal{L}_{VAE}||} + \frac{\mathcal{L}_{perc}}{||\nabla_x\mathcal{L}_{perc}||} + \frac{\mathcal{L}_{adv}}{||\nabla_x\mathcal{L}_{adv}||} \right],
\end{equation}
where $\mathcal{L}_{VAE}$ is the VAE loss, combining $l_2$ reconstruction loss and a Kullback-Leibler regularization term~\cite{kingma2022autoencoding}, the perceptual loss $\mathcal{L}_{perc}$ is based on LPIPS~\cite{zhang2018unreasonable}, while the adversarial loss $\mathcal{L}_{adv}$ employs a pre-trained VGG16~\cite{simonyan2015deep} discriminator with linear feature extraction and hinge loss. This VAE achieves a Wasserstein distance of 0.95 in the ZN reconstruction task.

\section{Original and Generated Histograms}
\label{sec:histograms}

\begin{figure}[t!]
    \centering
    \begin{subfigure}[t]{0.49\textwidth}
        \centering
        \includegraphics[width=\linewidth]{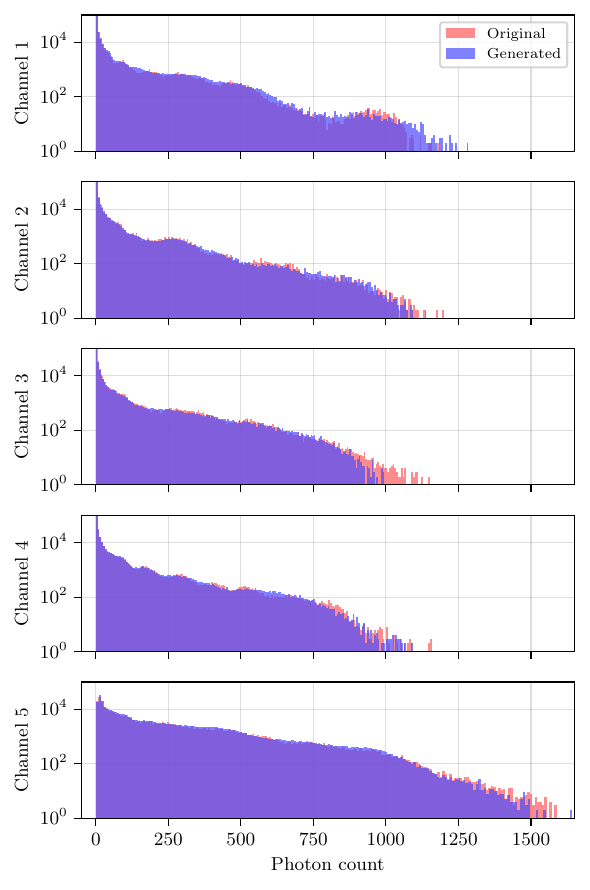}
        \caption{Histograms for the ZN detector.}
        \label{fig:histogram_n}
    \end{subfigure}
    \hfill
    \begin{subfigure}[t]{0.49\textwidth}
        \centering
        \includegraphics[width=\linewidth]{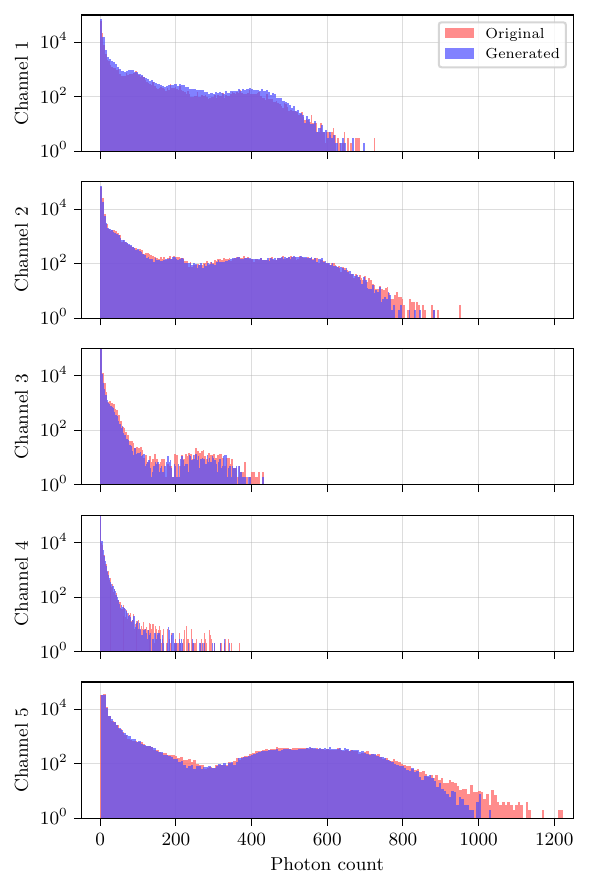}
        \caption{Histograms for the ZP detector.}
        \label{fig:histogram_p}
    \end{subfigure}
    \caption{Histograms depicting the sum of photons across the ZDC detector channels, comparing responses generated by GEANT (original) and FM (generated). The y-axis is presented on a logarithmic scale for better visualization of the data distribution.}
    \label{fig:histograms}
\end{figure}

In HEP simulations, it is essential that the simulated data accurately reflect underlying physical phenomena. This process commonly involves comparing histograms derived from simulation results against empirical data gathered from real instruments or, for fast simulations, against data from Monte Carlo simulations. Figure~\ref{fig:histograms}  shows the histograms from the five channels of the ZN and ZP responses, highlighting the fidelity of the FM model.

\section{Improved VQ-GAN}
\label{sec:vq_gan}

Building upon the findings of~\cite{wojnar2025fast}, we present an enhanced version of the VQ-GAN model~\cite{esser2021taming}, addressing the limited reconstruction performance (Figure~\ref{fig:all}, pink square). We focus on optimizing the first stage of training by reducing the downsampling factor from 8 to 4 and increasing the number of tokens from $6^2=36$ to $11^2=121$. Following~\cite{sun2024autoregressive}, we use a larger 512-entry codebook with reduced vector dimensionality of 8 and a smaller model of 69k parameters versus 1M. The training procedure and loss function remain consistent with Appendix~\ref{sec:setup}, resulting in a Wasserstein distance of 3.09 in the ZN reconstruction task. To enhance sampling quality, we use both temperature and top-k sampling in a generative pre-trained transformer (GPT) that serves as a learnable prior. We retain the same VQ-VAE model for the conditional variables as used in~\cite{wojnar2025fast}.

\begin{table}[t!]
    \small
    \centering
    \caption{Performance optimizations and their impact on the VQ-GAN model inference time (for batch size of 256). The last column highlights the improvement gained by applying optimizations sequentially, relative to the previous one.}
    \begin{tabular}{lcc}
        \toprule
        \textbf{Improvement}          & \textbf{Time [ms]} $\downarrow$ & \textbf{Relative change} \\
        \midrule
        Base model (from~\cite{wojnar2025fast}) & 0.09                 & --                       \\
        \specialrule{0.4pt}{0pt}{0pt}
        Greater number of steps (36 $\rightarrow$ 121) & 0.80                & +777\%                    \\
        Smaller model (4M $\rightarrow$ 1M params)     & 0.60                 & -25\%                    \\
        Mixed precision (F32 $\rightarrow$ F16)         & 0.45                 & -25\%                    \\
        \bottomrule
    \end{tabular}
	\label{tab:time_vqgan}
\end{table}

Table~\ref{tab:time_vqgan} summarizes GPT performance optimizations. The original model achieves an inference time of \SI{0.09}{\milli\second}. Increasing the number of steps leads to a substantial rise in time resulting in an inference time of \SI{0.80}{\milli\second} per sample. Reduction of the model size to 1 million parameters decreases the inference time by 25\%. Employed with mixed precision training and inference with the \textit{bfloat16} data type the final inference time is \SI{0.45}{\milli\second}.

\section{Direct Channel Value Estimation}
\label{sec:direct}

Since the electronics in ALICE capture only five discrete values, rather than a complete image of the particle shower, it may seem intuitive to directly estimate these values for each detector channel. However, this approach does not yield satisfactory results, and training an effective model proves to be a non-trivial challenge. Table~\ref{tab:direct} compares the performance of different models in the ZN and ZP simulation tasks. A densely connected neural network is trained using both standard regression and an FM generative scheme, following Appendix~\ref{sec:setup}. The classical models are implemented using the scikit-learn library~\cite{pedregosa2011scikit}, with all hyperparameters and source code accessible in the provided repository.

The results indicate that neural networks perform slightly better than classical ML methods; however, the FM model outperforms all other methods. This finding is particularly significant for the fast ZDC simulation, as it suggests that generating a full simulation and counting photons from the entire particle shower image is more effective than directly predicting only the five channel values.

\begin{table}[t!]
    \small
    \centering
    \caption{Performance comparison of direct channel estimation with full simulation using FM.}
    \begin{tabular}{lcc}
        \toprule
        \textbf{Model}      & \textbf{Wasserstein [ZN]} $\downarrow$ & \textbf{Wasserstein [ZP]} $\downarrow$ \\
        \midrule
        Original data       & 0.53  & 0.56  \\
        \specialrule{0.4pt}{0pt}{0pt}
        Linear regression   & 60.06 & 34.65 \\
        Gradient boosting   & 3.86  & 3.83  \\
        Decision tree       & 3.86  & 3.76  \\
        k-nearest neighbors & 2.98  & 3.48  \\
        Neural network      & 2.48  & 3.22  \\
        \specialrule{0.4pt}{0pt}{0pt}
        FM                  & \textbf{1.27} & \textbf{1.30} \\
        \bottomrule
    \end{tabular}
	\label{tab:direct}
\end{table}

\end{document}